\documentclass[journal,twoside,web]{ieeecolor}
\usepackage{generic}
\usepackage{cite}
\usepackage{amsmath,amssymb,amsfonts}
\usepackage{float}
\usepackage{algorithmic}
\usepackage{graphicx}
\usepackage{algorithm,algorithmic}
\usepackage{hyperref}
\hypersetup{hidelinks=true}
\usepackage{textcomp}
\def\BibTeX{{\rm B\kern-.05em{\sc i\kern-.025em b}\kern-.08em
		T\kern-.1667em\lower.7ex\hbox{E}\kern-.125emX}}
\usepackage{amsmath}
\usepackage{amssymb}

\usepackage{float}

\usepackage{caption}

\usepackage{multirow}

\usepackage{array}
\newcolumntype{C}{ >{\centering\arraybackslash} m{1.0cm} }
\newcolumntype{Q}{ >{\centering\arraybackslash} m{3 cmcm} }
\newcolumntype{M}[1]{>{\centering\arraybackslash}m{#1}}
\newcolumntype{P}[1]{>{\centering\arraybackslash}p{#1}}

\begin{document}

\title{TFBS-Finder: Deep Learning-based Model with DNABERT and Convolutional Networks to Predict Transcription Factor Binding Sites}
\author{Nimisha Ghosh,~\IEEEmembership{Member,~IEEE,}
        Pratik Dutta,
        and~Daniele Santoni
\thanks{N. Ghosh is with the Department
of Computer Science and Information Technology, Siksha `O' Anusandhan (Deemed to be University), Bhubaneswar, Odisha, India e-mail: (ghosh.nimisha@gmail.com).}
\thanks{P. Dutta is with Department of Computer Science and Engineering, Siksha `O' Anusandhan (Deemed to be University), Bhubaneswar, Odisha, India}
\thanks{D. Santoni is with Institute for System Analysis and Computer Science “Antonio Ruberti”, National Research Council of Italy, Rome, Italy}
\thanks{Manuscript received April 19, 2005; revised August 26, 2015.}}
\markboth{Journal of \LaTeX\ Class Files,~Vol.~14, No.~8, August~2015}%
{Ghosh \MakeLowercase{\textit{et al.}}: Bare Demo of IEEEtran.cls for IEEE Journals}

\maketitle

\begin{abstract}
Transcription factors are proteins that regulate the expression of genes by binding to specific genomic regions known as Transcription Factor Binding Sites (TFBSs), typically located in the promoter regions of those genes. Accurate prediction of these binding sites is essential for understanding the complex gene regulatory networks underlying various cellular functions. In this regard, many deep learning models have been developed for such prediction, but there is still scope of improvement. In this work, we have developed a deep learning model which uses pre-trained DNABERT, a Convolutional Neural Network (CNN) module, a Modified Convolutional Block Attention Module (MCBAM), a Multi-Scale Convolutions with Attention (MSCA) module and an output module. The pre-trained DNABERT is used for sequence embedding, thereby capturing the long-term dependencies in the DNA sequences while the CNN, MCBAM and MSCA modules are useful in extracting higher-order local features. TFBS-Finder is trained and tested on 165 ENCODE ChIP-seq datasets. We have also performed ablation studies as well as cross-cell line validations and comparisons with other models. The experimental results show the superiority of the proposed method in predicting TFBSs compared to the existing methodologies. The codes and the relevant datasets are publicly available at https://github.com/NimishaGhosh/TFBS-Finder/.
\end{abstract}


\section{Introduction}
Transcription factors (TFs) are proteins that play an important role in the regulation of genetic transcriptions by binding to DNA regulatory sequences known as Transcription Factor Binding Sites (TFBSs) (typically of size 4-30 bp)~\cite{Tompa2005, Tan2016, Qu2019}. Accurate prediction of TFBSs provides crucial information on regulatory networks that control cellular functions~\cite{LATCHMAN1997,Karin1990}, as well as identifying disease-associated variants and regulatory regions, enabling early detection and targeted therapies~\cite{Wang2024}. Chromatin immunoprecipitation sequencing (ChIP-seq)~\cite{Huang2022} can be used to identify DNA fragments that interact with TFs. However, such method is costly, paving the way for more inexpensive but effective computational approaches to predict TFBSs.

In recent years, many researchers have proposed several machine learning and deep learning approaches to identify TFBSs which include Hidden Markov Models (HMM)~\cite{Wong2013}, Support Vector Machine (SVM)~\cite{Djordjevic2003}, Random Forest (RF)~\cite{Xiao2009, Hooghe2012} models etc. Although traditional machine learning algorithms provide competitive results, they are dependent on other methods for feature extraction and they also do not work well with large-scale datasets. To address these challenges, recently many deep learning algorithms have been proposed, which show immense potential in predicting TFBSs when working with huge amount of experimental data such as ChIP-seq. In this regard, DeepSEA~\cite{Zhou2015} and DeepBind~\cite{Alipanahi2015} which use convolutional neural networks (CNNs) showcase good performance for the prediction of TFBSs. However, CNNs fail to capture long term dependencies between different positions in a DNA sequence. To overcome this, DanQ~\cite{Quang2016} proposed by Quang et al. and DeepSite~\cite{Zhang2020} proposed by Zhang et al., combine CNN and bidirectional long-short term memory (BiLSTM) to predict TFBSs. Other models such as DSAC~\cite{Yu2023}, D-SSCA~\cite{Zhang2021}, DeepSTF~\cite{Ding2023} and SAResNet~\cite{Shen2021} use attention mechanisms for TFBS prediction.

The evolutionary deep learning model Bidirectional Encoder Representations from Transformers (BERT)~\cite{devlin2019} effectively learns contextual information in natural language processing. Since its proposal in~\cite{devlin2019},  it has also been used for pre-training human DNA and protein sequences giving birth to DNABERT~\cite{Ji2021} and ProteinBERT~\cite{Brandes2022} respectively. 

DNABERT is an effective pre-trained model for embedding DNA sequences and subsequently other deep learning models may be used for classifying TFBSs. In this regard, Ghosh et al.~\cite{Ghosh2024} have used DNABERT for embedding the DNA sequences, while CNN, BiLSTM and capsule network have been used as the subsequent prediction layers. On the other hand, BERT-TFBS~\cite{Wang2024} has DNABERT-2, a CNN module, a convolution block attention module (CBAM) and an output module for TFBS prediction on 165 ChIP-seq datasets. Both works have shown promising results in predicting TFBSs. 

Taking cue from the literature, in this work we propose TFBS-Finder that utilises DNABERT, a CNN module, a modified convolution block attention module (MCBAM), a multi-scale convolutions with attention (MSCA) module and an output module to predict TFBSs. Similarly to~\cite{Wang2024}, we have trained and tested TFBS-Finder on 165 ChIP-Seq datasets, encompassing different cell-lines and TFs. The results show the superiority of TFBS-Finder when compared to the state-of-the-art predictors. 

To summarise, the main contributions of the work are as follows:
\begin{itemize}
    \item We propose a novel DNABERT-based deep learning model (TFBS-Finder) for predicting TFBSs. The model utilises DNABERT for embedding while CNN, MCBAM, MSCA and output modules are used for subsequent prediction.
    \item Ablation study is conducted to showcase the importance of each module of TFBS-Finder.
    \item To show the generalisability and robustness of TFBS-Finder, cross-cell line validations are conducted to predict TFBSs.
    \item All codes along with a sample dataset and a trained model are publicly available at https://github.com/NimishaGhosh/TFBS-Finder/..
\end{itemize}

\section{Materials and Methods}
In this section, data preparation is provided followed by a discussion on the pipeline of the work. 
\subsection{Data Preparation}
In this work, we have considered the benchmark 165 ChIP-seq datasets from Enclyopedia of DNA elements (ENCODE)~\cite{encode2012}, consisting of 29 different TFs from 32 cell lines. The dataset is taken from Zeng et al.~\cite{HZeng2016} where each positive sequence in the dataset is a 101 bp DNA sequence length containing TFBSs while the negative samples are obtained by shuffling the positive sequences conserving dinucleotide frequencies. Detailed descriptions of the datasets used in this work are given in Supplementary Table S1.

\subsection{Pipeline of the Work}
\begin{figure*}
		\centerline{
			\includegraphics[height=4.4in,width=7.2in]{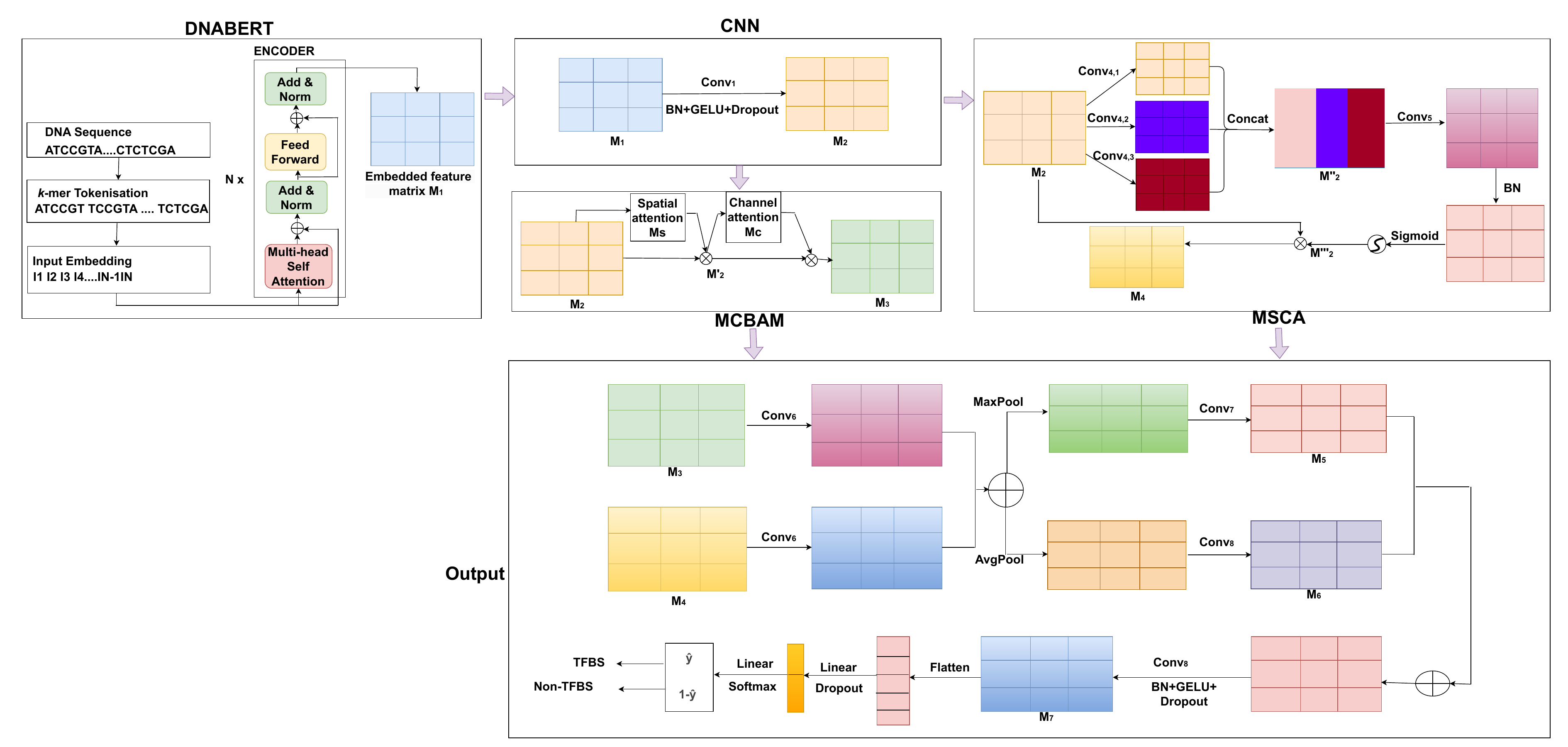}}
		\caption{Pipeline of the Work}
		\label{pipeline}
	\end{figure*}

The pipeline for the proposed work is given in Figure~\ref{pipeline}. The proposed model consists of five modules encompassing a DNABERT module, a Convolutional Neural (CNN) module, a modified CBAM (MCBAM) module, a multi-scale convolutions with attention module (MSCA) and an output module. 
\begin{itemize}
    \item DNABERT: DNABERT is pre-trained on DNA sequences to extract long-term dependencies within such sequences. Here, the sequences are divided into tokens of size $k$.
    \item CNN: DNABERT generates embeddings for $k$-mer tokens in DNA sequences, which capture context-aware representations of the input. The CNN module is used for extracting high-order local features from such embeddings by utilising a convolutional layer. 
    \item MCBAM: This module applies spatial and channel attention mechanisms to selectively enhance important local features extracted by the previous module. 
    \item MSCA: This module further enhances the local features captured by the feature extraction module. This is achieved by introducing multi-scale convolutions with an attention mechanism. Here, the attention is a form of feature-wise attention which is applied in the context of convolutional feature extraction.
    \item Output module: Using the relevant features captured by MCBAM and MSCA, this module uses such features and applies a series of convolutions, pooling and multi-layer perceptron to provide the final prediction of the presence or absence of TFBSs.
\end{itemize}
    \subsubsection{DNABERT}
    DNABERT is a pre-trained bidirectional encoder to enocode DNA sequences using $k$-mer technique. Sequences are divided into tokens of size $k$ and provided as initial input to DNABERT. Token and positional embeddings are applied to such tokens to form the input matrix \textbf{$M$}. This input matrix is then passed through $N$ sequential encoders to get the embedded matrix \textbf{$M_1$}. Each such encoder has multi-head self-attention, two layer of normalisations and a feed-forward network. The processing of the multi-head self attention mechanism on the input matrix for each encoder can be given as:
    \begin{equation}
    Multihead{(\textbf{M}^n)} = Concatenation(\textbf{head}_1^n,\dots,\textbf{head}_h^n)\mathcal{\textbf{W}}^{O,n}
		\label{bert1}
	\end{equation}
	where 
	\begin{multline}
		\textbf{head}_i^n = Softmax\left(\frac{{\textbf{M}^n}\mathcal{\textbf{W}}^{Q_i,n}({\textbf{M}^n}\mathcal{\textbf{W}}^{K_i,n})^T}{\sqrt{d_k}}\right).{\textbf{M}^n}\mathcal{\textbf{W}}^{V,n}_i
		\label{bert2}
	\end{multline}
	Here, all $\mathcal{W}$s are parameters learned during training of the model. Next, the residue connection $\textbf{M}^n$ and the multihead attention Multihead($\textbf{M}^n$) is passed through layer normalisation (LN) and feed forward network (FFN) to obtain:
    \begin{multline}
        \textbf{M}^{n+1} = LN(LN(\textbf{M}^n 
        +Multihead(\textbf{M}^n))+ \\
        FFN(LN(\textbf{M}^n+Multihead(\textbf{M}^n))))
    \end{multline}
    Finally, after passing through the $N$ sequential encoder layers, the embedding matrix \textbf{$M_1$} of size $d\times D$ is obtained where $d$ is the number of tokens while $D$ is the embedding dimension.

   \subsubsection{CNN Module}
   A single convolutional layer is used in this module to extract high-order local features from the embedded matrix \textbf{$M_1$}. This module consists of one convolutional block with convolutional operation $Conv_1$, batch normalisation (BN), Gaussian error linear unit 
   (GELU) activation function and dropout operation. This operations can be given as:
   \begin{equation}
       \textbf{M}_2 = Dropout(GELU(BN(Conv_1(\textbf{M}_1))))
       \label{CNN1}
   \end{equation}
\subsubsection{Modified Convolutional Block Attention Module}
\begin{figure*}
		\centerline{
			\includegraphics[height=0.7in,width=6.0in]{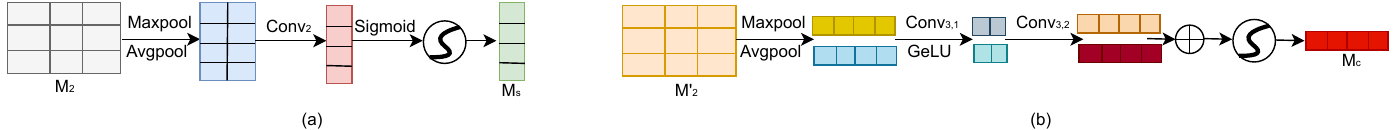}}
		\caption{Two submodules of MCBAM module where (a) represents the spatial attention block and (b) represents the channel attention block}
		\label{mcbam}
	\end{figure*}

   Spatial and channel attention blocks~\cite{Woo2018} are used in this module to enhance important local features obtained from the previous module. As given in~\cite{Woo2018}, the order of the two submodules affects the overall performance and in their work they have considered Channel-Spatial module. However, in our work (as reported later in Table~\ref{tab2}), spatial attention before channel attention provides a slightly better result. This placement of attention blocks has also shown better results in~\cite{Martinez2023}. The detailed operations of MCBAM module is given in Figure~\ref{mcbam}.
   
   As shown in Figures~\ref{pipeline} and~\ref{mcbam}, spatial attention is applied to feature matrix \textbf{$M_2$}. In this regard, initially \textbf{$M_2$} is subjected to global maxpooling and global average pooling in order to extract the spatial features. Convolution operation $Conv_2$ is then applied on such features and finally a sigmoid function is applied, resulting in spatial attention feature  \textbf{$M_S$}. Subsequently, element-wise matrix multiplication is carried out between \textbf{$M_2$} and \textbf{$M_S$} to generate \textbf{$M_2'$}.
   The aforementioned operations can be expressed as:
   \begin{equation}
       \textbf{M}_S= Sigmoid(Conv_2(Maxpool(\textbf{M}_2),Avgpool(\textbf{M}_2)))
       \label{sa}
   \end{equation}
   \begin{equation}
       \textbf{M}_2' = \textbf{M}_S \odot \textbf{M}_2
       \label{sa1}
   \end{equation}

   Once \textbf{$M_2'$} is constructed, channel attention submodule is applied on \textbf{$M_2'$} where each of its channel is subjected to global maxpooling and global average pooling separately. The resultant features are then individually passed through two convolutional layers where the first layer has convolution operation $Conv_{3,1}$ and a GELU activation function while the second layer consists of only a convolution operation $Conv_{3,2}$. The resultant feature channels are then added and passed through a sigmoid function to acquire the channel attention feature \textbf{$M_C$}. Element-wise matrix multiplication is then performed between \textbf{$M_C$} and \textbf{$M_2'$} resulting in the feature matrix \textbf{$M_3$} which is the final output of MCBAM module. These operations can be given as:
   \begin{multline}
   \textbf{M}_C = Sigmoid(Conv_{3,2}(GELU(Conv_{3,1}(Maxpool(\textbf{M}_2')))\\ 
+Conv_{3,2}(GELU(Conv_{3,1}(Avgpool(\textbf{M}_2')))) 
       \label{ca}
   \end{multline}
   \begin{equation}
       \textbf{M}_3 = \textbf{M}_C \odot \textbf{M}_2'
       \label{ca1}
   \end{equation}
   
\subsubsection{Multi-Scale Convolution with Attention Module}
This module utilises multi-scale convolutions with attention to further enhance the local features captured by the feature extraction module. As shown in Figure~\ref{pipeline},
three separate convolutions $Conv_{4,1}$, $Conv_{4,2}$ and $Conv_{4,3}$ are applied to each input channel of the feature matrix \textbf{$M_2$}. The results of such convolutions are concatenated along the channel dimension producing \textbf{$M_2''$}, thus retaining all multi-scale information across channels. Next, a point-wise convolution $Conv_{5}$ is applied on \textbf{$M_2''$}, thereby combining and compressing the concatenated features into a unified representation, enabling the model to learn meaningful relationships between features from different scales. Subsequently, a batch normalisation and a sigmoid activation function are applied to capture attention weights. Finally, element-wise multiplication is performed between such attention weight matrix \textbf{$M_2'''$} and \textbf{$M_2$} to get the final output. In this way, features that are deemed more relevant (with higher weights) are enhanced, while less relevant features are suppressed. Such operations can be summarised as:
\begin{equation}
    \textbf{M}_2'' = Concat(Conv_{4,i}(\textbf{M}_2)), \qquad i\in\{1,2,3\}
    \end{equation}
    \begin{equation}
    \textbf{M}_2''' = Sigmoid(BN(Conv_{5}(\textbf{M}_2'')))
    \end{equation}
    \begin{equation}
    \textbf{M}_4 = \textbf{M}_2 \odot \textbf{M}_2'''
\end{equation}
Thus, \textbf{$M_4$} is the final matrix of the MSCA module.
\subsubsection{Output Module}
The output block considers the feature matrices \textbf{$M_3$} and \textbf{$M_4$} respectively, obtained from MCBAM and MSCA modules. This has the effect of parallel attention~\cite{Martinez2023} in the output module, thereby exploiting both MCBAM and MSCA modules. The matrices obtained from these modules are passed separately through convolutional operations $Conv_{6}$ and the resultant matrices are combined via addition which are then passed in parallel through global maxpooling and average pooling separately. The resultant matrices \textbf{$M_5$} and \textbf{$M_6$} are then concatenated and passed through yet another convolution operation with GELU as the activation function along with batch normalisation and dropout operations, leading to the feature matrix \textbf{$M_7$}. Finally, the prediction is obtained by flattening \textbf{$M_7$} and subsequently applying multilayer perceptron which consists of a fully connected layer with dropout as well as another fully connected layer with softmax activation function. The output of this module is the prediction probability $\hat{y}$, which predicts if a given DNA sequence consists of TFBSs. All the aforementioned operations can be depicted as:
\begin{multline}
  \textbf{M}_5, \textbf{M}_6 = Conv_{7}(Maxpool(Conv_{7}(\textbf{M}_3)+Conv_{6}(\textbf{M}_4)),\\ 
  Avgpool(Conv_{6}(\textbf{M}_3)+Conv_{7}(\textbf{M}_4))) 
\end{multline}
\begin{equation}
    \textbf{M}_7 = Dropout(GELU(BN(Conv_8(\textbf{M}_5+\textbf{M}_6))))
\end{equation}
\begin{equation}
    \hat{y} = Softmax(Linear(Dropout(Linear(Flatten(\textbf{M}_7)))))
\end{equation}

\subsection{Model Training}
For training purpose, cross-entropy loss function is used in this work~\cite{Wang2024}:
\begin{equation}
    Loss (y,\hat{y}) = -\frac{1}{\eta}\sum_{i=1}^\eta(y_i(log\hat{y}) + (1-y_i)log(1-\hat{y_i}))
\end{equation}
Here, $y$ are the actual values of the DNA sequences while $\hat{y}$ are the predicted values of TFBS-Finder and $\eta$ represents the batch size of DNA sequences which is considered as 64 in this work. AdamW optimisation method is used in this work while ReduceLROnPlateau is used as the scheduler which automatically aids in reduction of learning rate when the value of area under the precision-recall curve plateaus. Moreover, to prevent overfitting, early stopping and dropout are also used in this work where the dropout values are set to 0.2 and 0.3. The training is performed for 15 epochs. All the hyperparameters are selected based on experiments. All such hyperparameter values are provided in Supplementary Tables S2 and S3.

\section{Results and Discussions}
This section discusses the different results as obtained in this work. Initially, the analysis is performed to check if TFBS-Finder is able to differentiate sequences containing TFBS from shuffled sequences. As TFBS-Finder model is made up of a combination of 5 modules, we have conducted ablation studies to evaluate the contribution of such modules. We have also investigated the ability of TFBS-Finder when trained on sequences coming from a given cell-line to recognise sequences of different cell-lines, observing substantial cross- cell line consistency. Subsequently, we have performed comparison studies with existing state-of-the-art predictors to show the superiority of TFBS-Finder.

\subsection{Performance Metrics}
The different metrics used in this work to showcase the prediction performance of TFBS-Finder are accuracy, PR-AUC and ROC-AUC. Accuracy represents the proportion of correctly predicted samples. However, accuracy can turn out to be a biased metric when the data is imbalanced. In this regard, PR-AUC and ROC-AUC are some other parameters to judge prediction performance. PR-AUC is the area under the precision-recall curve and represents the model's overall performance. Its value lies between 0 and 1 where a higher value of PR-AUC depicts better prediction performance. For imbalance class distributions, PR-AUC is a better metric than Accuracy. Another metric which is equally suitable for evaluating predictor performance is ROC-AUC or Area Under the Receiver Operating Characteristic curve, which is a graphical representation of the performance of a binary classification model at various classification thresholds. In order to evaluate the performance of TFBS-Finder, all the three metrics have been used in this work. However, since early stopping is used in this work to avoid overfitting, PR-AUC is used as the early stopping criteria. If PR-AUC does not improve for 2 consecutive runs during validation, the model stops training. We have used different values of $k$ (3, 4, 5 and 6) and observed that with $k$=5, we get the best results. Thus, all the following experiments are conducted with $k$ = 5. 


\subsection{Ablation Study}

TFBS-Finder uses different important modules such as DNABERT, CNN, MCBAM and MSCA for predicting TFBSs. In order to showcase their individual contributions, experiments and ablation studies are performed on 165 ChIP-seq datasets with different variants to compare such variants. 
\begin{itemize}
\item DNABERT+Output (Variant 1): In this variant, only DNABERT+Output is considered. All other modules are removed. 
    \item DNABERT+CNN+Output (Variant 2): Only DNABERT with CNN and output modules are used in this variant. 
    \item DNABERT+CNN+MSCA+Output (Variant 3): In this variant, the MCBAM module is removed. 
    \item DNABERT+CNN+MCBAM+Output (Variant 4): In this variant, the MSCA module is removed. 
    
    \item DNABERT+CNN+CBAM+MSCA+Output (Variant 5): Here, the positions of spatial and channel attention modules are reversed (such position is similar to CBAM~\cite{Woo2018}) that is, channel attention comes before spatial attention module in order to show the importance of their individual position in the MCBAM module. 

\end{itemize}

\begin{table}[H]
    \centering
    \begin{tabular}{lccc}
Model	&	Accuracy	&	PR-AUC	&	ROC-AUC	\\\hline
Variant 1	&	0.918	&	0.954	&	0.955	\\
Variant 2	&	0.925	&	0.957	&	0.956	\\
Variant 3	&	0.900	&	0.942	&	0.942	\\
Variant 4	&	0.909	&	0.950	&	0.950	\\
Variant 5	&	0.927	&	0.961	&	0.961	\\
TFBS-Finder	&	\textbf{0.930}	&	\textbf{0.961}	&	\textbf{0.961}	\\\hline
\end{tabular}
\caption{Experimental study for TFBS-Finder and its five variant models based on average values of accuracy, PR-AUC and ROC-AUC on 165 ChIP-seq datasets.}
\label{tab2}
\end{table}
As observed from the results in Table~\ref{tab2} in Variant 1, with accuracy, PR-AUC and ROC-AUC scores of 0.918, 0.954 and 0.955, DNABERT provides strong representations of both local motifs and global dependencies. It performs well because the self-attention mechanism is inherently capable of capturing the relationships between tokens at various scales. The addition of CNN in Variant 2 captures localized features or motifs in the DNA sequences that may complement the global representations from DNABERT, thereby providing an accuracy of 0.925 and PR-AUC and ROC-AUC of 0.957 and 0.956 respectively. In variant 3, MSCA introduces a mechanism to expand the receptive field of CNNs, which helps capture broader local context providing an accuracy, PR-AUC and ROC-AUC score of 0.900, 0.942 and 0.942 respectively. However, DNABERT already excels at capturing global context through self-attention, so adding only MSCA might create redundancy or disrupt DNABERT’s well-calibrated global representations. Spatial and channel attention mechanisms in Variant 4 focus on specific regions of feature maps and refines the features captured by the CNN module leading to scores of 0.909, 0.950 and 0.950 for all the parameters.
Although these mechanisms can improve certain tasks, they might conflict with global attention of DNABERT by overemphasizing local features, leading to a loss of global context. Significant improvement (accuracy, PR-AUC and ROC-AUC values of 0.927, 0.961 and 0.961) is noticed for Variant 5 when MSCA is added to channel and spatial attention mechanisms. The proposed model TFBS-Finder shows the best performance across all variants in which an optimal balance is observed between local and global feature extraction. Here, CNN captures localised patterns while spatial and channel attention in MCBAM module refine these patterns by focusing on relevant regions and channels. The MSCA module also provides an additional mechanism to capture multi-scale features and broader local context using dilated convolutions. This increases the receptive field and allows the model to learn a diverse set of features at various scales. All these modules together achieve an accuracy, PR-AUC and ROC-AUC of 0.930, 0.961 and 0.961 respectively. 
The detailed results are provided in Supplementary Table S4.
\begin{figure*}
		\centerline{
			\includegraphics[height=2.5in,width=2.5in]{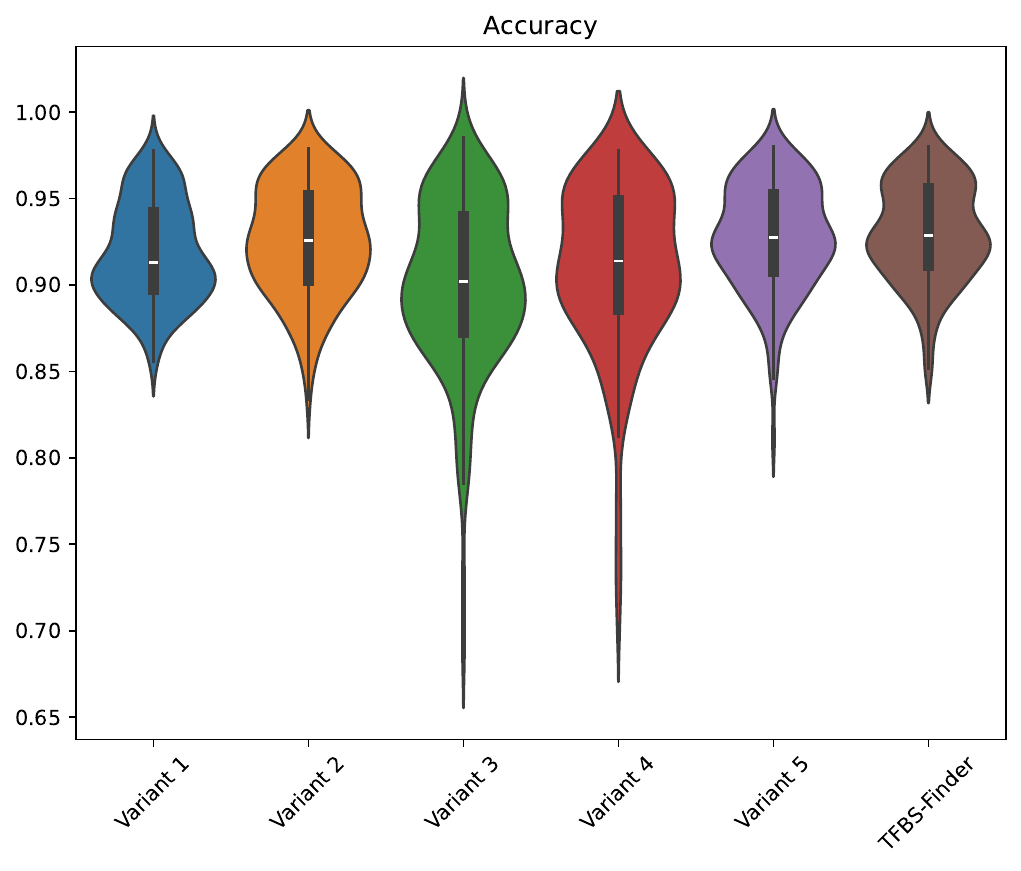}
            \includegraphics[height=2.5in,width=2.5in]{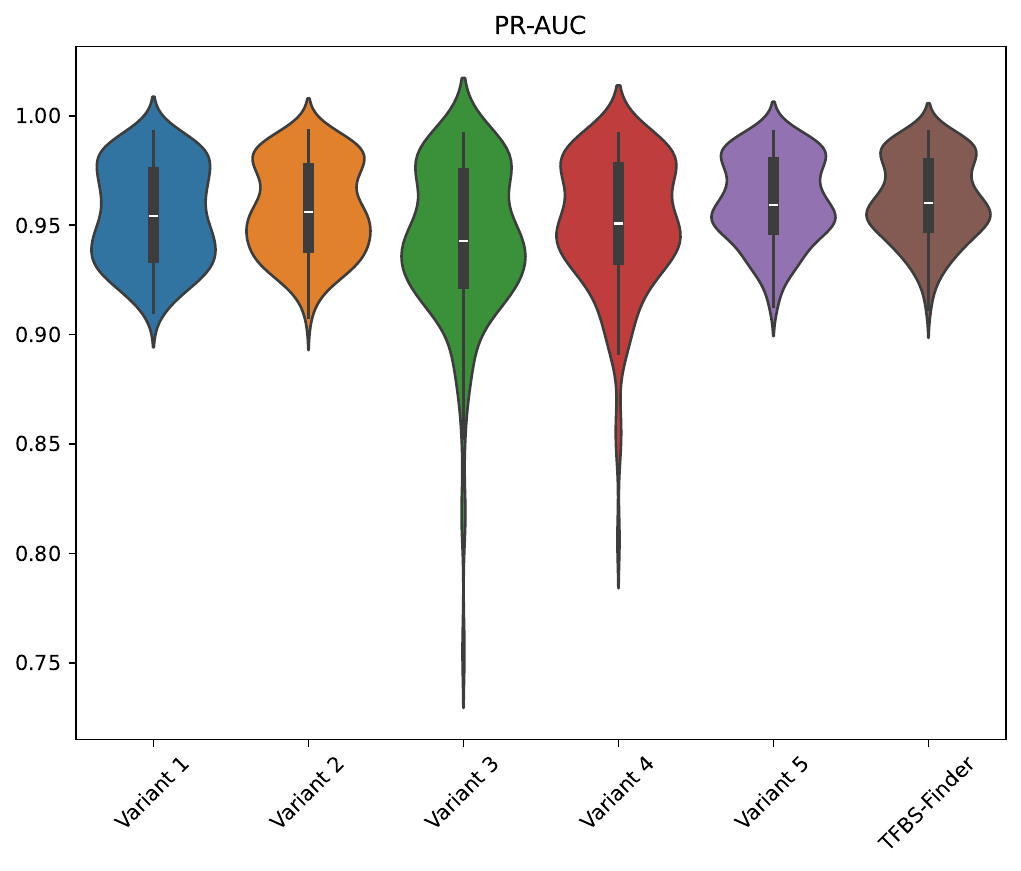}
            \includegraphics[height=2.5in,width=2.5in]{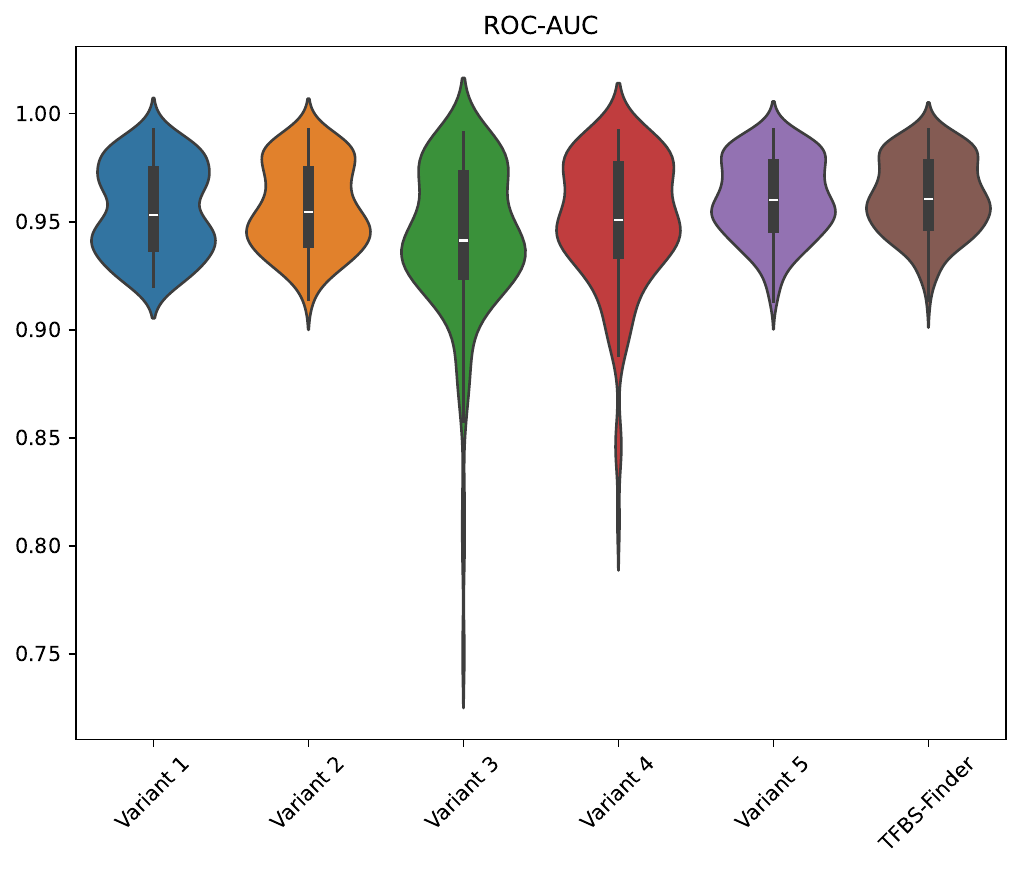}}
            \centerline{(a)\hspace{56mm}(b)\hspace{56mm}(c)}
		\caption{Prediction performance of TFBS-Finder compared with other variants where (a), (b) and (c) represents accuracy, PR-AUC and ROC-AUC. The white line inside the violins represent the median while the bold black lines show the interquartile range. The two vertical thin lines inside the violins show the range of maximum and minimum non-outlier values and the very large thin lines represent the outliers.}
		\label{comparison_variant}
	\end{figure*}

Figure~\ref{comparison_variant} shows the violin plots for the prediction performance of TFBS-Finder and the 5 variants. The median values of accuracy for the 5 variants and TFBS-Finder are respectively 0.913, 0.925, 0.902, 0.913, 0.927 and 0.928 while for PR-AUC (and ROC-AUC respectively), such values are 0.954, 0.956, 0.942, 0.950, 0.959 and 0.960 (0.953, 0.954, 0.941, 0.950, 0.959, 0.960). In terms of quartile values (lower and upper) TFBS-Finder also shows superior performance. Such values for lower (and upper) quartiles for TFBS-Finder in terms of accuracy, PR-AUC and ROC-AUC are respectively 0.911, 0.948, 0.948 (0.955, 0.978, 0.976). Among the variants, Variant 5 has the closest performance to TFBS-Finder. The lower (and upper) quartiles for Variant 5 for accuracy, PR-AUC and ROC-AUC are 0.907, 0.948, 0.947 (0.952, 0.978, 0.976). Considering all the metrics, it can be observed that although Variant 5 gives equally competent results, the experiments show that TFBS-Finder works slightly better for all the performance metrics. 

\subsection{Cross-cell line validation}
\begin{figure*}
		\centerline{
			\includegraphics[height=1.2in,width=1.5in]{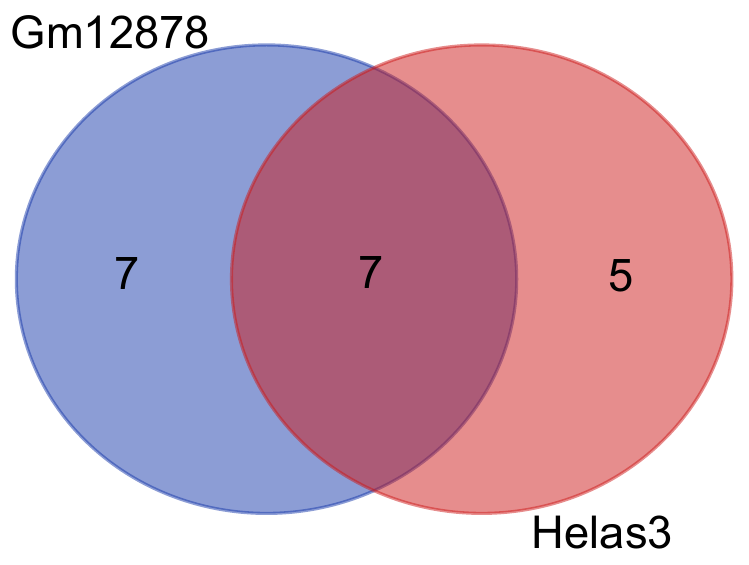}
            \includegraphics[height=1.2in,width=1.5in]{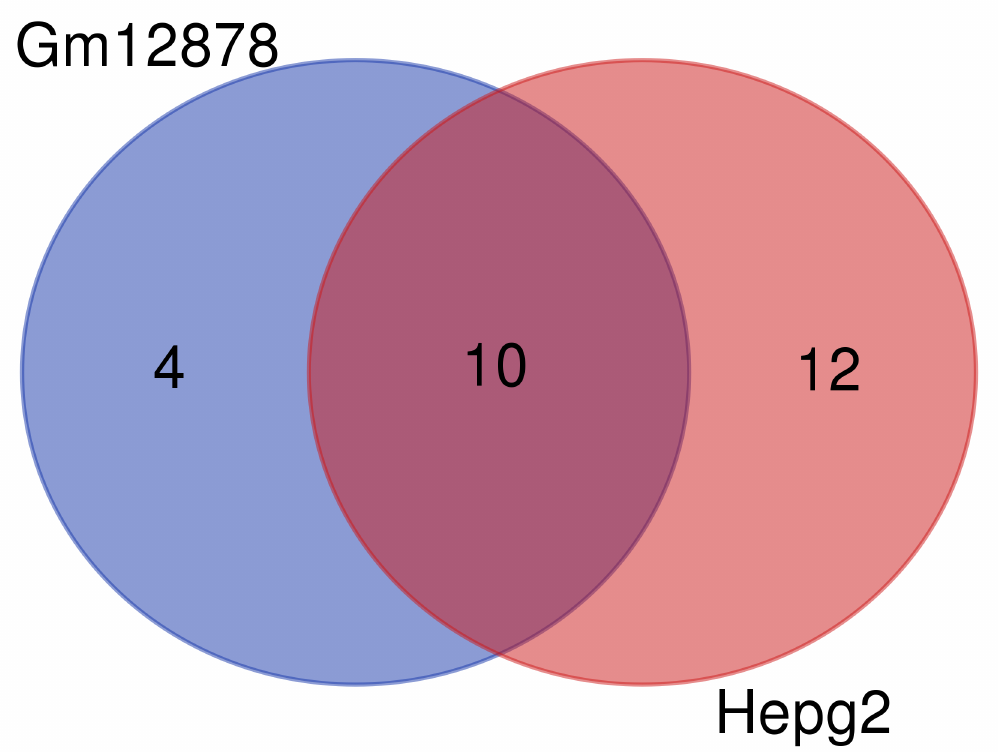}
            \includegraphics[height=1.2in,width=1.5in]{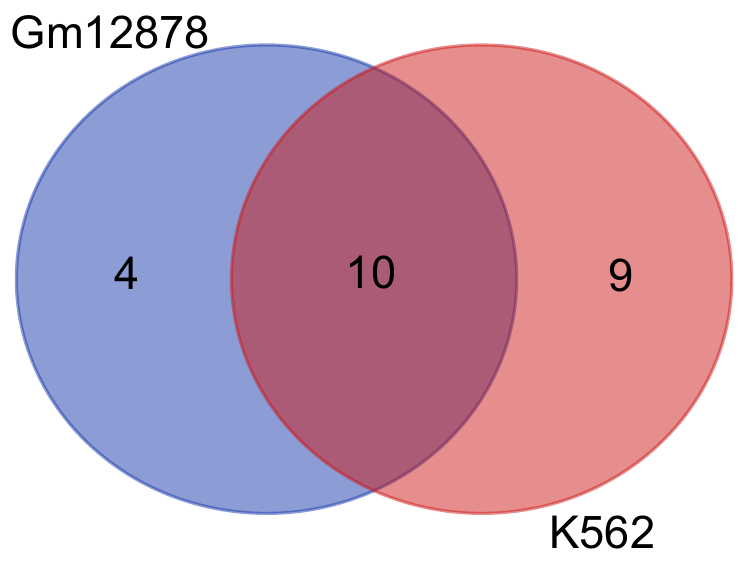}}
            \centerline{(a)\hspace{35mm}(b)\hspace{35mm}(c)}
		
        \centerline{
			\includegraphics[height=1.2in,width=1.5in]{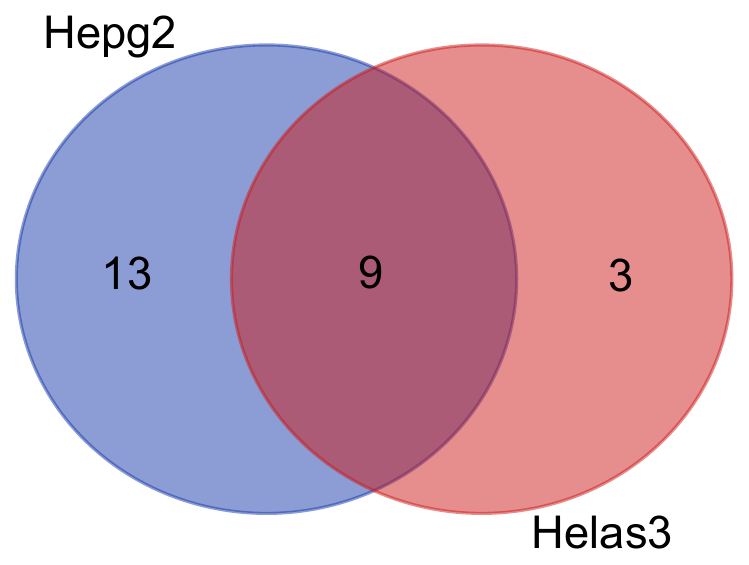}
            \includegraphics[height=1.2in,width=1.5in]{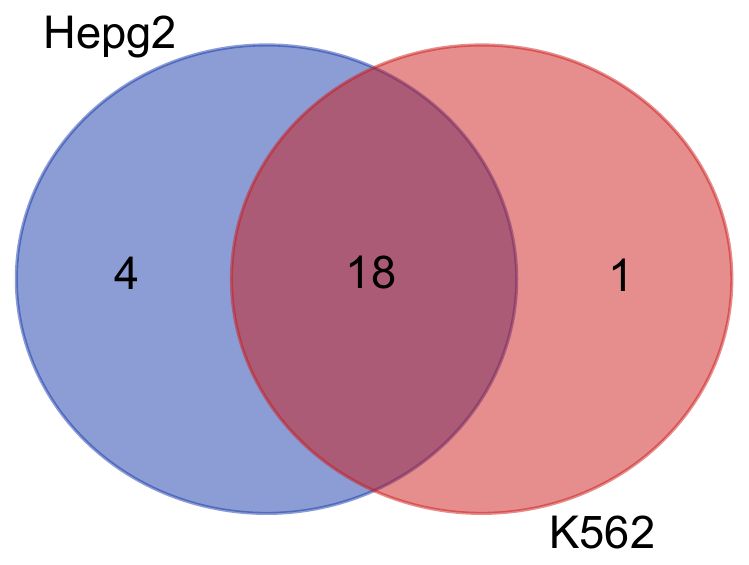}
            \includegraphics[height=1.2in,width=1.5in]{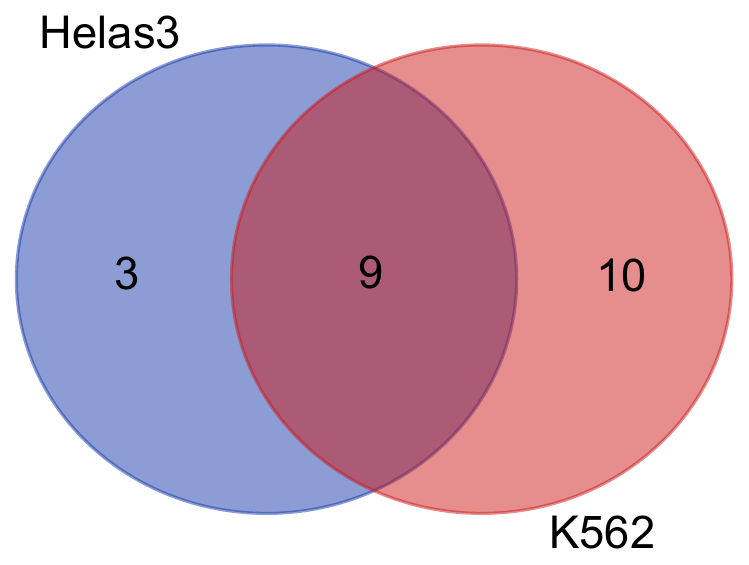}}
            \centerline{(d)\hspace{35mm}(e)\hspace{35mm}(f)}
		\caption{Venn Diagram to show the common TFs between (a) Gm12878 and Helas3, (b) Gm12878 and Hepg2, (c) Gm12878 and K562, (d) Hepg2 and Helas3, (e) Hepg2 and K562 and (f) Helas3 and K562}
		\label{venn}
	\end{figure*}

Cross-cell line validation is performed in order to show the generalisability and robustness of TFBS-Finder to identify the binding sites of a particular TF, in this case CTCF, occurring in different cell lines. In this regard, we have considered four cell lines Gm12878, Helas3, Hepg2 and K562 which have many TFs in common (as shown in Figure~\ref{venn})  and also have a large number of CTCF sequences. 

To prepare the training dataset for cross-cell validation, the training sequences bounded by CTCF are considered as positive while the rest of the sequences bounded by other TFs are considered as negative. This resulted in 71761 positive and 147870 negative sequences for Gm12878. Similarly, for Helas3, these statistics are 84901 and 166311 while for Hepg2 such numbers of positive and negative sequences are 35857 and 381333 and for K562, they are 82297 and 406752.
As can be observed from the statistics, the number of negative sequences is quite high as compared to the positive sequences which is quite obvious given the fact that only the sequences bounded by CTCF forms the positive sequences while those  bounded by the other TFs together constitute the negative sequences. As the motivation is to identify the binding sites of CTCF, to balance the dataset we have randomly selected a number of negative sequences equal to the number of positive ones while ensuring that the negative sequences have candidates from all the other TFs (71761, 84901, 35857 and 82297 for Gm12878, Helas3, Hepg2 and K562 respectively). The test dataset is also prepared in a similar manner, where the sequences  bounded by CTCF are considered as positive and those  bounded by other TFs are negative. For example, for CTCF pertaining to Gm12878 (for wgEncodeAwgTfbsBroadGm12878CtcfUniPk), the number of positive sequences in test set is 8644 indicating that there are 8644 such sequences in the dataset wgEncodeAwgTfbsBroadGm12878CtcfUniPk with TFBSs for CTCF while 37098 negative sequences constitute the other 15 TFs. For the other TFs, the number of positive sequences includes the binding sequences of CTCF in wgEncodeAwgTfbsBroadGm12878CtcfUniPk and wgEncodeAwgTfbsUtaGm12878CtcfUniPk, while the negative sequences are their own binding sites. This method of data preparation is also true for the other cell lines. Detailed statistics is provided in Supplementary Table S5. 

Table~\ref{crosscell} shows the average values over all the TFs of ROC-AUC scores for TFBS-Finder trained on the four cell lines Gm12878, Helas3, Hepg2 and K562 and then tested on the testing samples of these cell lines to predict the corresponding CTCF binding sites. The diagonal values indicate the traditional cell line prediction while the rest of the values show the cross-line validation results. It can be observed from the results that TFBS-Finder has a high score of more than 0.93 in all the cases, thereby showcasing the generalisability and robustness of the proposed model. 
\begin{table}
    \centering
    \begin{tabular}{llcccc}\hline
    \multicolumn{6}{c}{Testing}\\\hline
&Cell lines	&	Gm12878	&	Helas3	&	Hepg2 & K562	\\\hline
\multirow{4}{*}{\rotatebox{90}{Training}}&Gm12878	&	0.941	&	0.949	&	0.954 & 0.925	\\
&Helas3	&	0.937	&	0.957	&	0.961 & 0.936	\\
&Hepg2	&	0.935	&	0.952	&	0.957 & 0.926	\\
&K562	&	0.941	&	0.959	&	0.957 & 0.936	\\\hline
\end{tabular}
\caption{Average scores for cross-cell line and traditional validations based on ROC-AUC values for predicting CTCF binding sites. Cross-cell line validations indicate that TFBS-Finder is trained on a cell line and is then used to predict CTCF binding sites in other cell lines.}
\label{crosscell}
\end{table}

\begin{figure*}
		\centerline{
			\includegraphics[height=2.5in,width=3.5in]{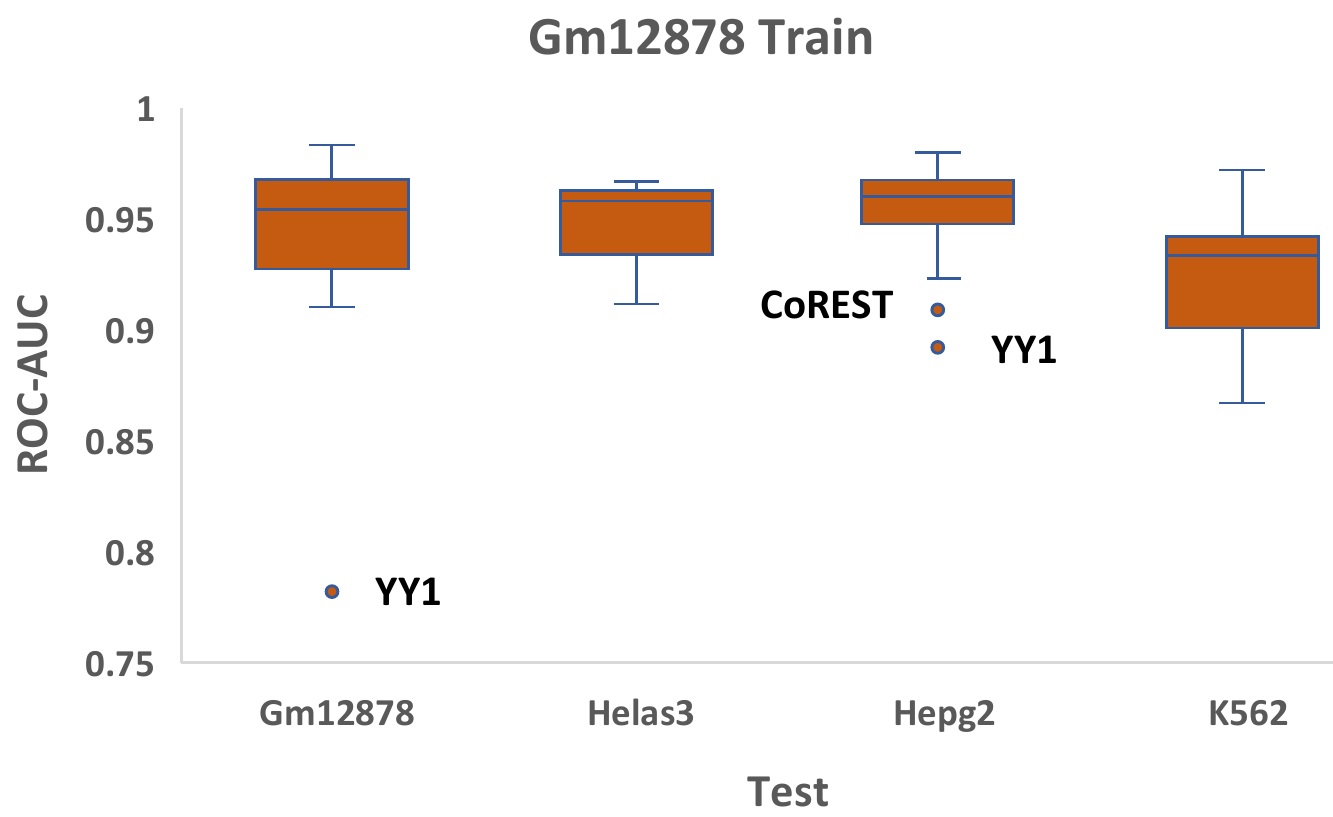}
            \includegraphics[height=2.5in,width=3.5in]{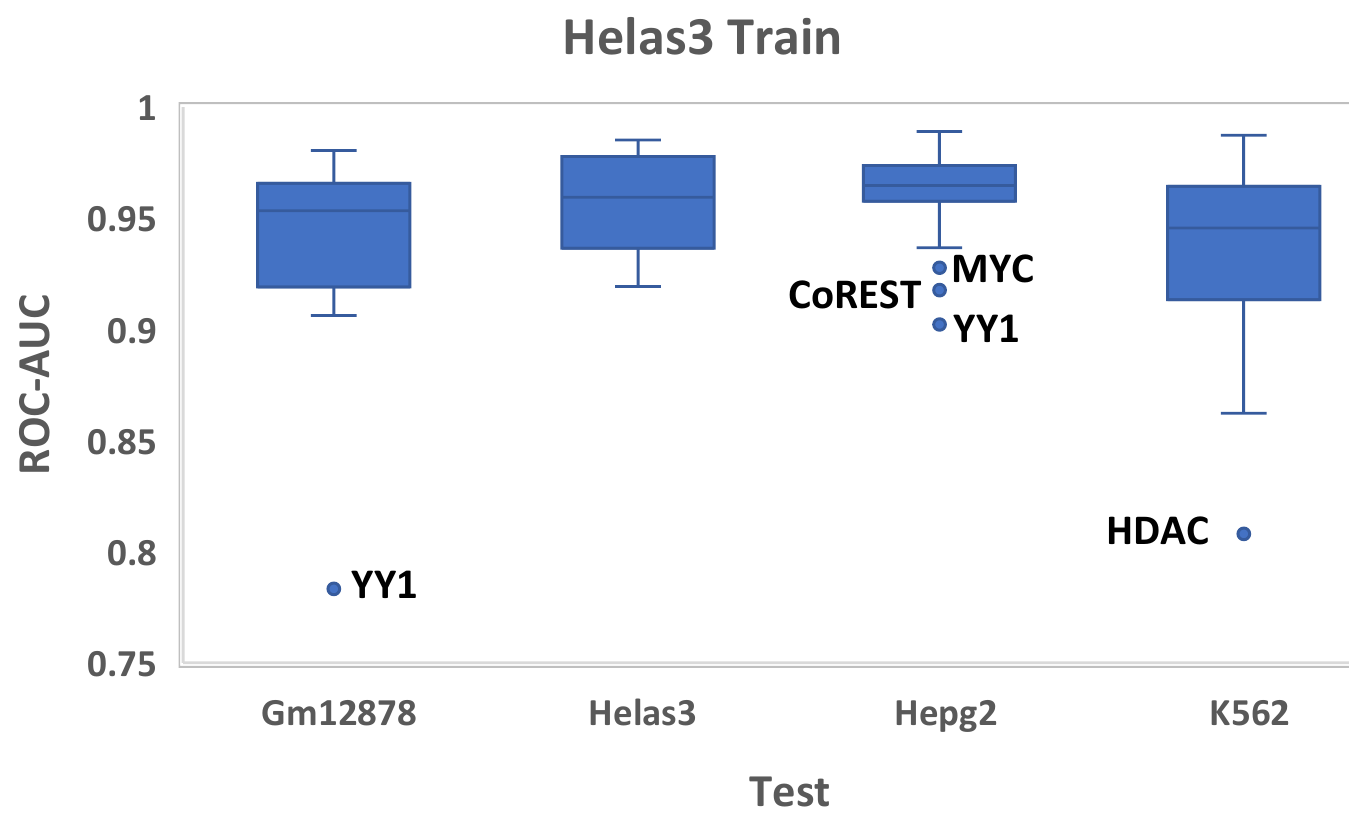}}
            \centerline{(a)\hspace{56mm}(b)}
            \centerline{
            \includegraphics[height=2.5in,width=3.5in]{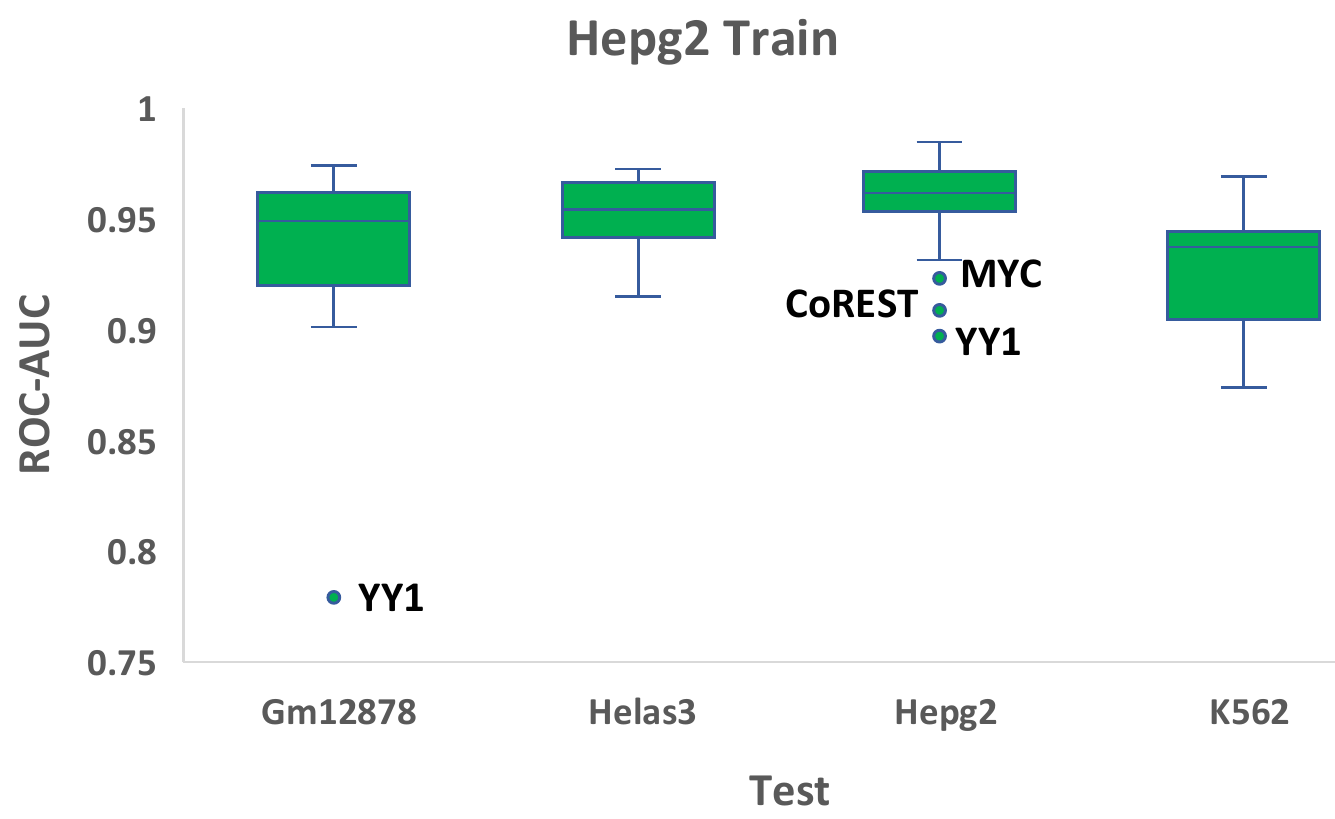}
            \includegraphics[height=2.5in,width=3.5in]{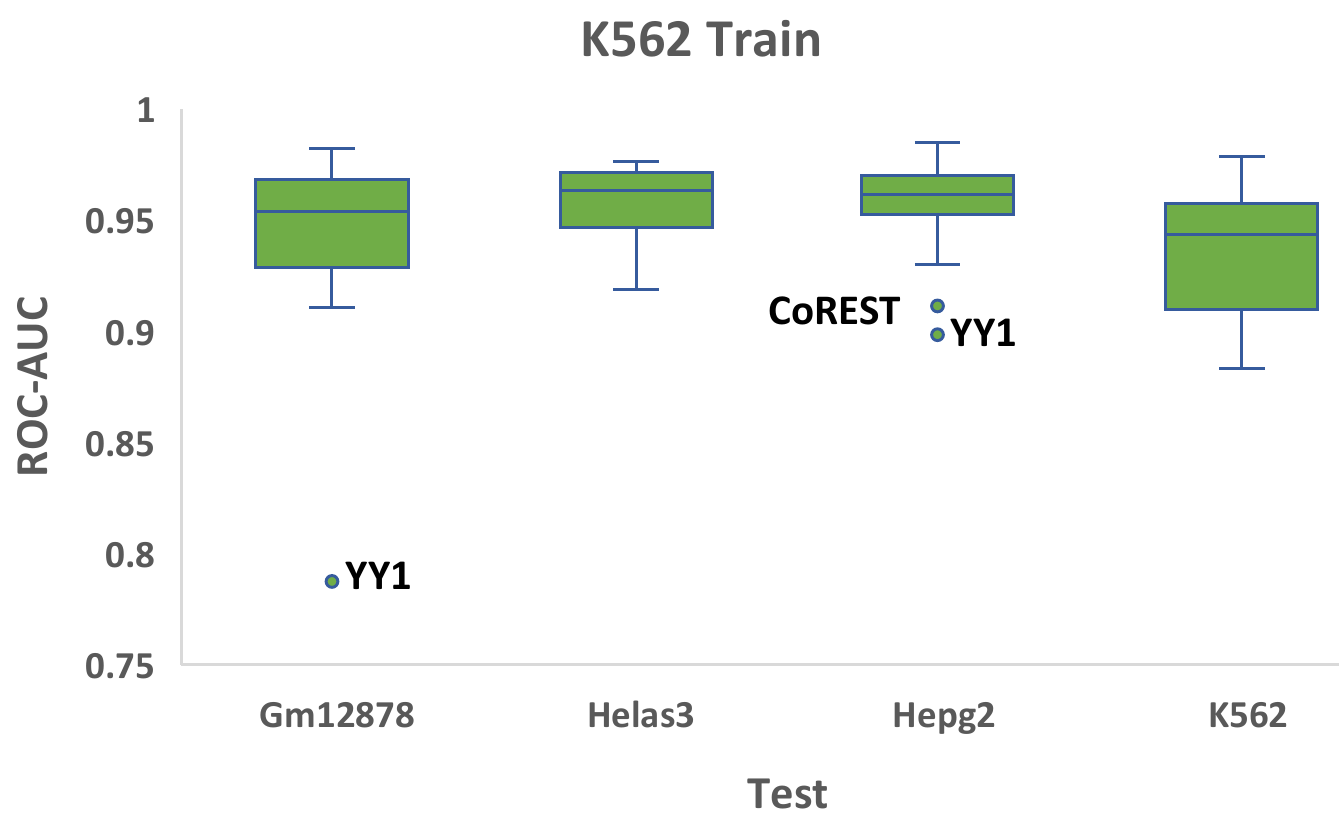}}
            \centerline{(c)\hspace{56mm}(d)}
		\caption{Prediction performance of TFBS-Finder for cross-cell line and traditional validations where (a), (b), (c) and (d) show the results for training on Gm12878, Helas3, Hepg2 and K562. The line inside the boxes represent the median while the top and bottom edges show the upper and lower quartiles respectively. The circles represent the outlier values.}
		\label{cross}
	\end{figure*}
Figure~\ref{cross} shows the distributions of the prediction performance of TFBS-Finder based on ROC-AUC when trained on Gm12878, Helas3, Hepg2 and K562 while tested for the related cell lines. For Hepg2, the traditional validation shows superior performance with the median value as 0.961. However, for the other cell-lines, cross-validations show slightly better performance. As shown in~\cite{Wang2024}, CTCF and YY1 show similarity in their binding regions, thereby often misclassifying sequences bound by YY1 as CTCF. This is also observed in all the figures of Figure~\ref{cross}. Moreover, CoREST is also misclassified as CTCF for all the cell lines while MYC is misclassified in Helas3 and Hepg2 while HDAC is misclassified in Helas3. This show a probable interaction of CoREST, MYC and HDAC with CTCF. To summarise, the results of Table~\ref{crosscell} and Figure~\ref{cross} indicate that TFBS-Finder shows good performance for both cross-cell line and traditional validations. The detailed results are provided in the Supplementary Table S6.

\subsection{Comparison with existing models}
In this subsection, we present the results of the comparison of the prediction performance of TFBS-Finder with other existing models. In this regard, we have considered seven benchmark models which include BERT-TFBS~\cite{Wang2024}, DeepBind~\cite{Alipanahi2015}, DanQ~\cite{Quang2016}, DLBSS~\cite{ZhangQ2021}, CRPTS~\cite{WANG2021}, D-SSCA~\cite{Zhang2021} and DSAC~\cite{Yu2023}. BERT-TFBS uses DNABERT-2~\cite{zhou2024dnabert2} as the encoding technique along with CNN and CBAM modules while DeepBind considers one-hot encoding along with CNN architecture. On the other hand, DanQ utilises CNN as well as Bi-LSTM architectures, DLBSS and CRPTS use a shared CNN and CNN with RNN architectures, respectively, while D-SSCA and DSAC use CNNs and attention mechanisms. The results of this comparison are based on 165 ChIP-seq datasets where the dataset is the same as considered in~\cite{Wang2024}. 
\begin{table}[H]
    \centering
    \begin{tabular}{lccc}
Model	&	Accuracy	&	PR-AUC	&	ROC-AUC	\\\hline
DanQ	&	0.782	&	0.855	&	0.849	\\
 DeepBind 	&	0.785	&	0.858	&	0.853	\\
CRPTS	&	0.793	&	0.867	&	0.862	\\
DLBSS	&	0.793	&	0.871	&	0.865	\\
D-SSCA	&	0.793	&	0.871	&	0.867	\\
DSAC	&	0.816	&	0.891	&	0.887	\\
BERT-TFBS	&	0.851	&	0.920	&	0.919	\\
TFBS-Finder	&	\textbf{0.930}	&	\textbf{0.961}	&	\textbf{0.961}	\\\hline
\end{tabular}
\caption{Comparison among TFBS-Finder and other state-of-the-art predictors considering the  verage values of accuracy, PR-AUC and ROC-AUC.}
\label{tab3}
\end{table}
Table~\ref{tab3} shows the comparison among TFBS-Finder and other existing predictors considering the average values of accuracy, PR-AUC and ROC-AUC over all cell lines and all TFs. As can be observed from the table, TFBS-Finder outperforms all other models. The best performance among the existing predictors is shown by BERT-TFBS. However, TFBS-Finder with accuracy, PR-AUC and ROC-AUC of 0.930, 0.961 and 0.961 respectively, outperforms BERT-TFBS by 7.9\%, 4.1\% and 4.2\%. The detailed results are provided in the Supplementary Table S7.
\begin{figure*}
		\centerline{
			\includegraphics[height=2.5in,width=2.5in]{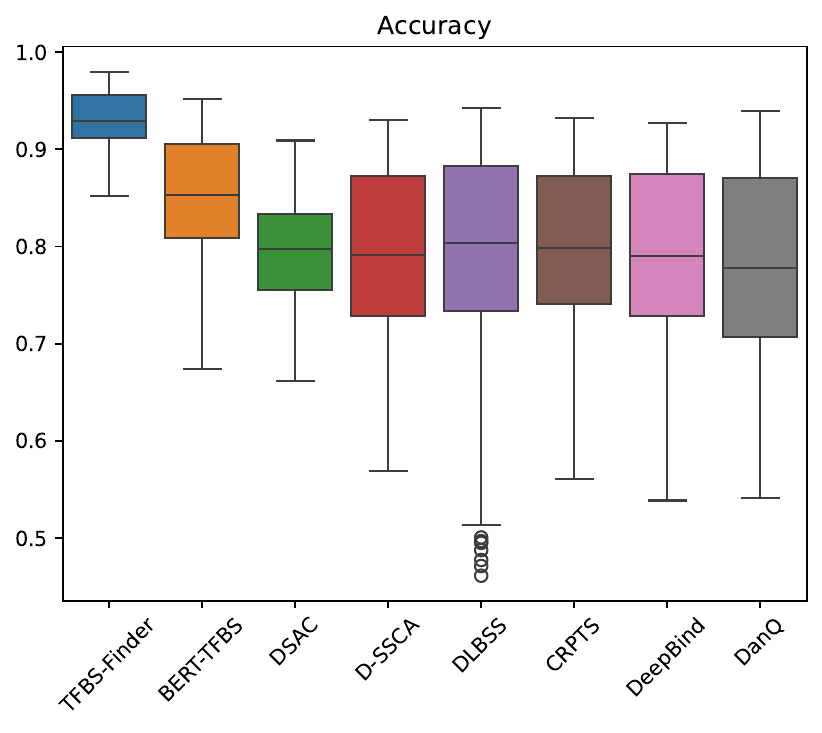}
            \includegraphics[height=2.5in,width=2.5in]{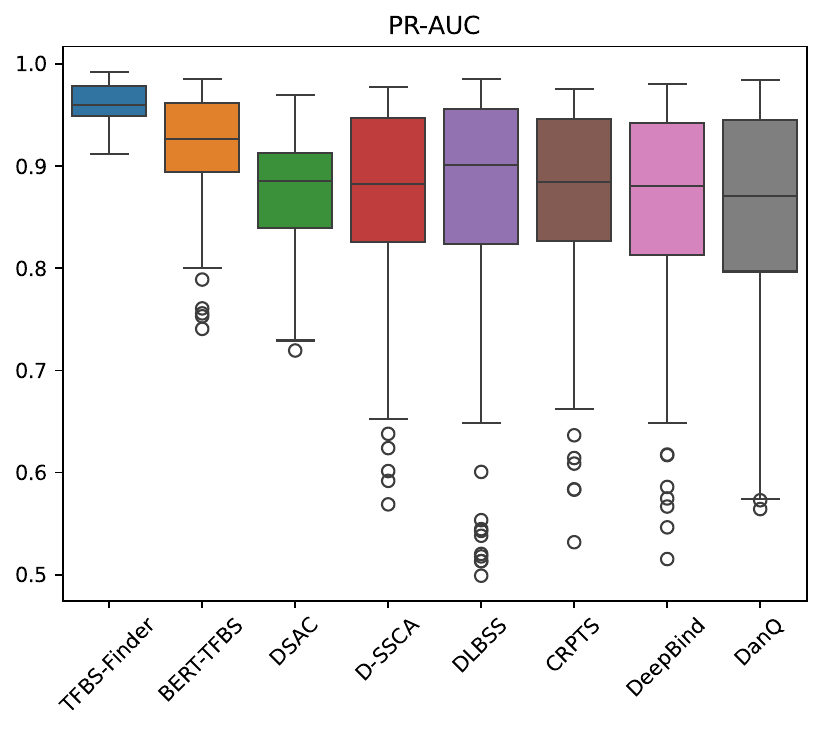}
            \includegraphics[height=2.5in,width=2.5in]{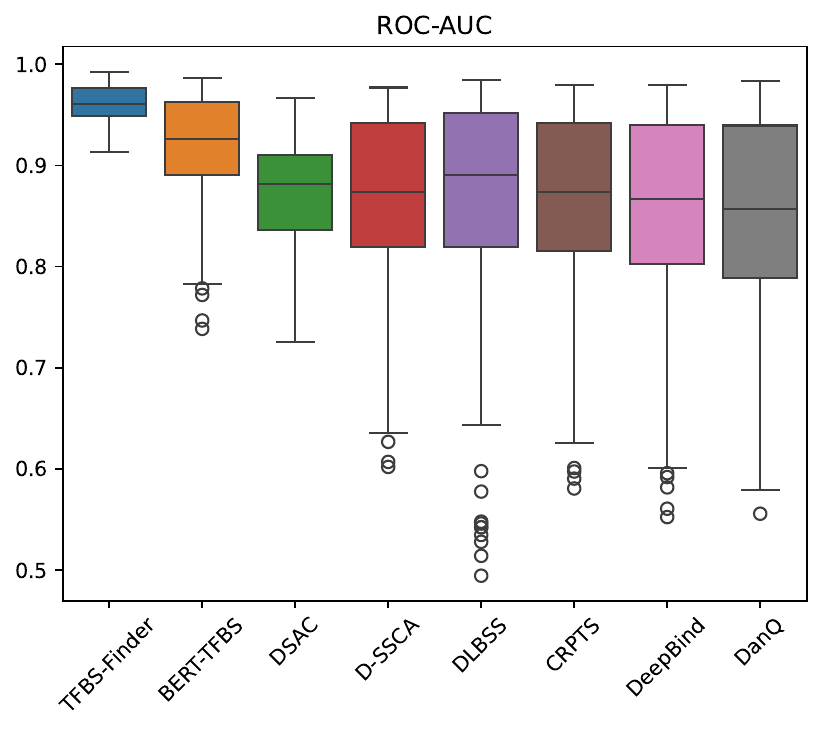}}
            \centerline{(a)\hspace{56mm}(b)\hspace{56mm}(c)}
		\caption{Prediction performance of TFBS-Finder compared with other state-of-the-art predictors where (a), (b) and (c) represents accuracy, PR-AUC and ROC-AUC. The line inside the boxes represent the median while the top and bottom edges show the upper and lower quartiles respectively. The circles represent the outlier values.}
		\label{comparison}
	\end{figure*}

Figure~\ref{comparison} reports the distributions of accuracy, PR-AUC and ROC-AUC scores of TFBS-Finder along with other state-of-the-predictors. TFBS-Finder shows the best performance amongst all the benchmark predictors in terms of median, lower and upper quartiles. TFBS-Finder has median values of 0.928, 0.960 and 0.960 for accuracy, PR-AUC and ROC-AUC which are  7.5\%, 3.4\%  and 3.4\% better than BERT-TFBS. The lower quartiles (and upper quartiles) values for accuracy, PR-AUC and ROC-AUC scores of TFBS-Finder are 0.911, 0.948 and 0.948 respectively (0.955, 0.978 and 0.976)
which are better than BERT-TFBS by  10.3\%, 5.5\% and 5.8\%(1.6\%, 5\% and 1.3\%). The above results showcase the superiority of TFBS-Finder over other existing state-of-the-art predictors in predicting TFBSs on the 165 ChIP-seq datasets.

\section{Conclusion and Future Work}
In this work, we propose TFBS-Finder which utilises DNABERT for global context while CNN, MCBAM and MSCA performs feature extraction and capture local contexts to predict TFBSs for DNA sequences. TFBS-Finder is trained and tested on 165 ChIP-seq datasets where the experimental studies show that TFBS-Finder outperforms the existing predictors in terms of average accuracy, PR-AUC and ROC-AUC. We have performed ablation studies to understand the importance of local and global contexts to identify TFBSs. Furthermore, cross-cell line validations are also performed to show the generalisability and robustness of TFBS-Finder. Although, the proposed model shows very inspiring results, as a future work, we would additionally like to incorporate DNA structure information along with DNA sequences to see the effect on the prediction capacity of TFBS-Finder. 

\section{Acknowledgment}
This work has been carried out under SERB-SURE scheme (SUR/2022/002836), funded by Government of India. The authors show their gratitude to the Italian CINECA Consortium for graciously providing the required computational resources.
\bibliographystyle{IEEEtran}
\bibliography{ref}
\end{document}